\documentclass[journal,twoside,web]{ieeecolor}
\usepackage[table,xcdraw]{xcolor}
\usepackage{jsen}
\usepackage{subfigure}
\usepackage{graphicx}
\usepackage{amsmath,amssymb,amsfonts}
\usepackage{wrapfig}
\usepackage{multirow}
\usepackage{hyperref}
\hypersetup{
    colorlinks=true,
    linkcolor=red,
    anchorcolor=black,
    citecolor=blue,
    filecolor=cyan,
    menucolor=red,
    runcolor=cyan,
    urlcolor=blue
}
\usepackage{orcidlink}

\graphicspath{{Figures/}}

\def\BibTeX{{\rm B\kern-.05em{\sc i\kern-.025em b}\kern-.08em
    T\kern-.1667em\lower.7ex\hbox{E}\kern-.125emX}}
\definecolor{abstractbg}{rgb}{0.89804,0.94510,0.83137}
\begin{document}
\title{STAL: Spike Threshold Adaptive Learning Encoder for Classification of Pain-Related Biosignal Data}
\author{Freek~Hens~\orcidlink{0009-0005-0405-0560}, 
Mohammad Mahdi~Dehshibi~\orcidlink{0000-0001-8112-5419}~\IEEEmembership{Senior Member, IEEE},
Leila~Bagheriye~\orcidlink{0000-0003-1605-5850}~\IEEEmembership{Member, IEEE}, 
Mahyar~Shahsavari~\orcidlink{0000-0001-7703-6835},~\IEEEmembership{Member, IEEE},
Ana~Tajadura-Jim{\'e}nez~\orcidlink{0000-0003-3166-3512}
\thanks{The paper is submitted for review on \today. ``This work was supported by the European Union's Horizon 2020 program under grant No 101002711 for the BODYinTRANSIT project.'' }
\thanks{Freek~Hens (e-mail: freek.hens@outlook.com), Mahyar~Shahsavari, and Leila~Bagheriye are with the Donders Institute for Brain, Cognition and Behavior at Radboud University in Nijmegen, The Netherlands. }
\thanks{Mohammad Mahdi~Dehshibi and Ana~Tajadura-Jim{\'e}nez are with the Department of Computer Science and Engineering at Universidad Carlos III de Madrid, L{\'e}ganes, Spain.}}

\IEEEtitleabstractindextext{%
\fcolorbox{abstractbg}{abstractbg}{%
\begin{minipage}{\textwidth}%
\begin{wrapfigure}[12]{r}{3in}%
\vspace{-0.25cm}%
\hspace*{-0.25cm}\includegraphics[width=3in]{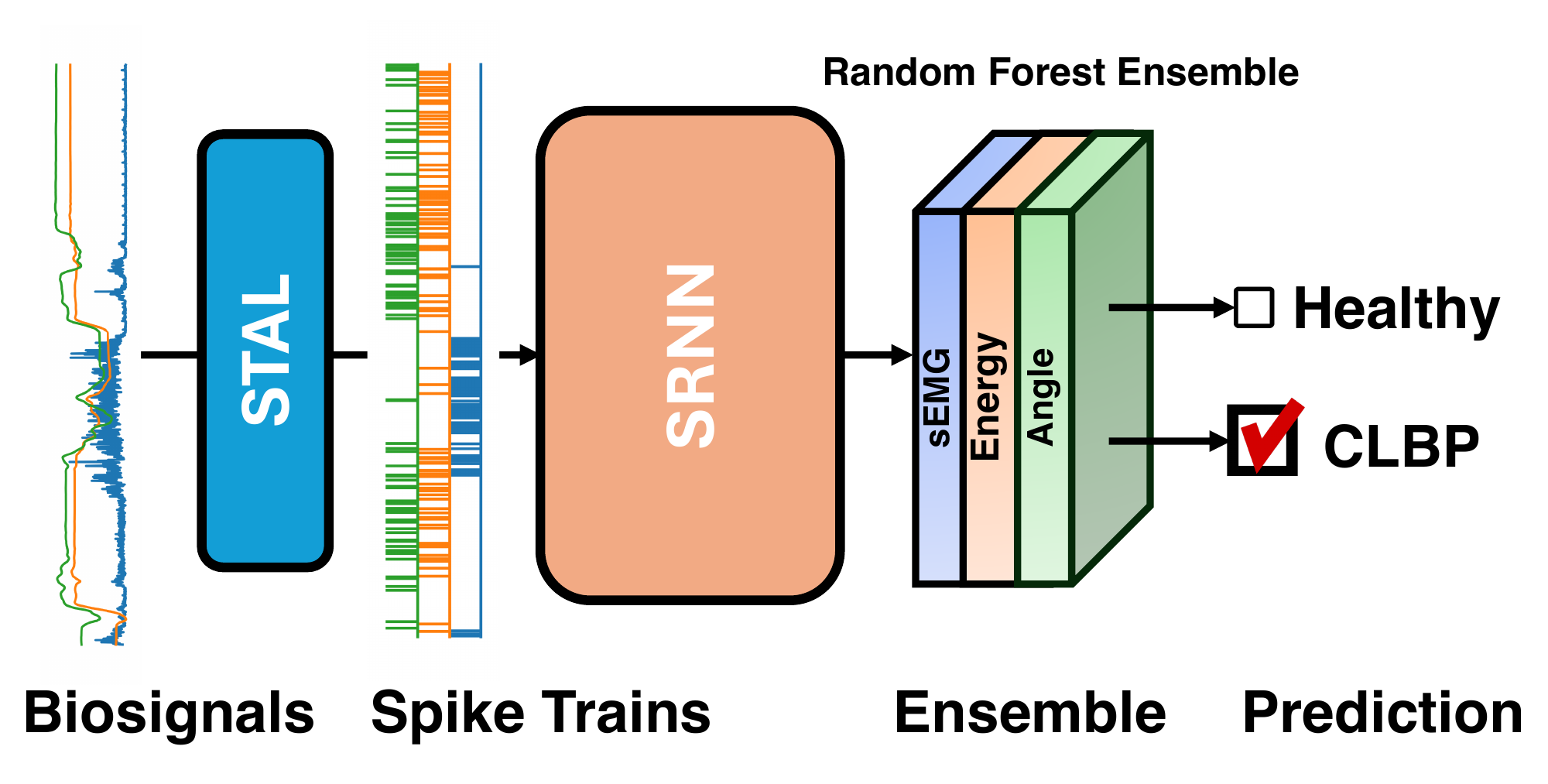}%

\end{wrapfigure}%
\begin{abstract}
This paper presents the first application of spiking neural networks (SNNs) for the classification of chronic lower back pain (CLBP) using the EmoPain dataset. Our work has two main contributions. We introduce Spike Threshold Adaptive Learning (STAL), a trainable encoder that effectively converts continuous biosignals into spike trains. Additionally, we propose an ensemble of Spiking Recurrent Neural Network (SRNN) classifiers for the multi-stream processing of sEMG and IMU data. To tackle the challenges of small sample size and class imbalance, we implement minority over-sampling with weighted sample replacement during batch creation. Our method achieves outstanding performance with an accuracy of 80.43\%, AUC of 67.90\%, F1 score of 52.60\%, and Matthews Correlation Coefficient (MCC) of 0.437, surpassing traditional rate-based and latency-based encoding methods. The STAL encoder shows superior performance in preserving temporal dynamics and adapting to signal characteristics. Importantly, our approach (STAL-SRNN) outperforms the best deep learning method in terms of MCC, indicating better balanced class prediction. This research contributes to the development of neuromorphic computing for biosignal analysis. It holds promise for energy-efficient, wearable solutions in chronic pain management.
\end{abstract}

\begin{IEEEkeywords}
Spike Train Encoder, Spiking Recurrent Neural Network, EmoPain, Chronic Pain
\end{IEEEkeywords}
\end{minipage}}}

\maketitle
\section{Introduction} 
\IEEEPARstart{B}{iosignals} are pivotal in non-invasively analysing and understanding physiological processes and human body function~\cite{supratak2016survey}. Among these, surface electromyography (sEMG) and data from inertial measurement units (IMUs) offer a comprehensive representation of an individual's motor behaviour~\cite{li2024noninvasive}. sEMG measures the electrical activity of skeletal muscles, while IMUs directly quantify body segment movements. Analysing human motor behaviour has found widespread applications in several domains, particularly in chronic pain management~\cite{aung2016EmoPain}.

Chronic lower back pain (CLBP) is a challenging condition that impacts millions of individuals worldwide, often resulting in reduced quality of life and significant economic strain. The precise detection and monitoring of CLBP are essential for developing effective pain management strategies~\cite{yong2022prevalence}. However, the analysis of biosignals for this purpose presents several challenges. One primary challenge is signal contamination from various sources, including motion artefacts and crosstalk from nearby physiological processes~\cite{fernandez2023systematic}. Another lies in inter- and intra-individual variability due to physiological differences or factors like fatigue and stress~\cite{makaram2021analysis}. Beyond these challenges, sEMG data are high-dimensional and multi-channel, with a non-stationary nature due to variations in muscle anatomy between individuals. Similarly, IMU signals present challenges such as drift and integration errors, sensitivity to sensor placement, and the complexity of interpreting 3D motion data~\cite{dehshibi2023pain}.

Conventionally, researchers have addressed these challenges using handcrafted features and filtering techniques. While promising, these approaches often require domain expertise and may fail to capture the intricate dynamics of complex biosignals~\cite{farago2023review}. More recently, deep learning models such as convolutional and recurrent neural networks~\cite{jorge2024LSFAN,dehshibi2023pain} have demonstrated the potential to derive intricate features from raw data for CLBP classification. However, these models typically demand large, high-quality datasets for training and are computationally intensive.

Spiking Neural Networks (SNNs) offer a promising alternative for biosignal analysis. Inspired by biological neural systems, SNNs process information through discrete spike events, mimicking the communication mechanism of biological neurons. This biological plausibility is particularly relevant for sEMG and IMU analysis, as SNNs can capture the complex temporal dynamics of these signals more effectively~\cite{wang2019acquisition}. Moreover, when deployed on neuromorphic hardware, SNNs can operate with significantly lower power consumption compared to other neural network architectures. This feature makes them ideal for resource-constrained environments like wearable devices~\cite{sun2023feasibility}.

Despite these advantages, a significant challenge in using SNNs for biosignal analysis lies in converting continuous analogue signals into binary spike trains -- the native language of SNNs. Various encoding methods have been proposed, including rate-based, time-to-first-spike, and delta-modulated approaches. However, these methods often struggle to preserve the temporal dynamics of biosignals, resist inherent noise, and adapt to signal characteristics without violating biological constraints~\cite{auge2021reviewSNN,perez2021sparse}.

In this paper, we present the first application of neuromorphic computing techniques to CLBP classification using sEMG and IMU data provided in the EmoPain dataset~\cite{aung2016EmoPain}. This dataset includes recordings from 25 healthy individuals and 21 subjects with CLBP performing a sequence of physical exercises. We introduce two key innovations:

\begin{enumerate}
\item Spike Threshold Adaptive Learning (STAL): A learnable encoder designed to convert continuous sEMG and IMU signals into spike trains that efficiently preserve temporal information.
\item An ensemble of Spiking Recurrent Neural Network (SRNN) classifiers: This multi-stream SRNN uses recurrent leaky integrate-and-fire neurons to process the spike trains and improves accuracy and robustness in classifying individuals as healthy or having CLBP with an ensemble of Random Forest classifiers.
\end{enumerate}

By combining the biological plausibility and efficiency of SNNs with the proven performance of ensemble learning methods, our two-fold contribution provides opportunities for developing more effective, personalised rehabilitation strategies for individuals with CLBP. Furthermore, the proposed approach has potential broader impacts beyond CLBP classification. It could be adapted for other biosignal analysis tasks in healthcare, such as detecting neurodegenerative disorders or monitoring cardiovascular health. The low-power consumption of SNNs also makes this approach promising for long-term, continuous health monitoring using wearable devices.

The remainder of this paper is organised as follows: Section~\ref{sec:related_work} reviews related work in biosignal encoding and studies that addressed challenges defined in the EmoPain database. Section~\ref{sec:proposed_system} details our proposed method, including the STAL encoder and the ensemble of SRNN classifiers. Section~\ref{sec:experiments_results} presents our experimental setup and results, comparing our method's performance to existing approaches. Finally, Section~\ref{sec:conclusion} concludes the paper and discusses future research directions.

\section{Literature Review} \label{sec:related_work}

The use of spiking neural networks in biosignal processing has shown great promise, thanks to their biological plausibility, lower power consumption, and reduced training requirements compared to conventional deep learning methods~\cite{SNN_POETS}. However, their potential application in chronic pain detection and identification of pain-related protective behaviour is an unexplored area, presenting a unique opportunity in this domain.

\begin{figure*}[!htbp]
    \centering
    \includegraphics[width=\linewidth]{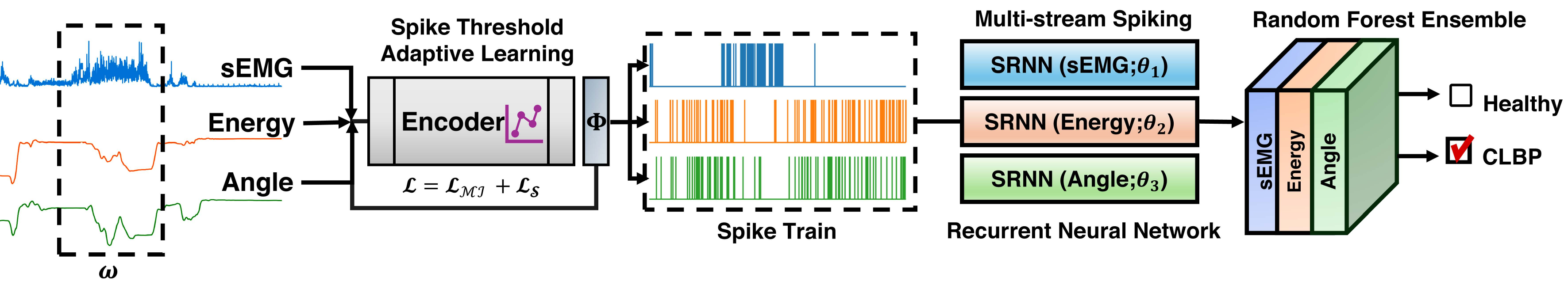}
    \caption{A schematic of the proposed architecture, including the Spike Threshold Adaptive Learning (STAL) encoder and Spiking Recurrent Neural Network (SRNN), which are responsible for classifying data from EmoPain~\cite{aung2016EmoPain} into Healthy and Chronic Lower Back Pain (CLBP) classes. The multimodal biosignals consist of sEMG sensors and derived features (i.e., Joint Energy and Joint Angle) from IMU sensors. These continuous signals convert into spike trains using the STAL encoder and are subsequently classified using an ensemble of SRNNs. The ensemble prediction is generated using a Random Forest meta classifier.}
    \label{fig:arch-overview}
\end{figure*}

A critical component of using SNNs in biosignal processing is the conversion of continuous signals into discrete spike trains, known as spike encoding. This process is particularly challenging for EMG and IMU data due to their rapidly changing patterns and susceptibility to noise~\cite{dehshibi2023pain}. Several encoding methods have been proposed, each with distinct trade-offs in temporal resolution, noise robustness, and computational efficiency.

The use of rate-based encoding, though energy-efficient, often faces challenges in capturing the rapid changes inherent in EMG and IMU signals~\cite{fabian2023sparse}. Conversely, phase coding provides noise resilience at the cost of complex synchronisation mechanisms~\cite{ARRAY2023}. Other techniques, including Send-on-Delta, Ben's spiker algorithm~\cite{eshraghian2023snntorch}, and Variational mode decomposition (VMD)~\cite{donati2019discrimination}, have their unique advantages in precise spike generation or direct encoding of analogue signals but frequently encounter difficulties in handling the intricate dynamics of \emph{in vivo} EMG and IMU data and fail to fully leverage the spatial information in high-density EMG arrays.

Recent research has explored learnable encoders for multimodal biosignal analysis in SNNs~\cite{gong2023spiking}, offering adaptability to specific signal characteristics. However, these approaches often need more focus on high temporal resolution, noise robustness, and computational efficiency, which are crucial for real-world CLBP applications using wearable sensors~\cite{gurchiek2021wearables}.

The limitations of existing encoding methods in handling the unique challenges of EMG and IMU data for CLBP classification motivate the development of a new approach. An ideal encoder for this purpose should retain temporal dynamics, resist inherent noise, and adjust to signal characteristics without violating biological constraints. Hence, we introduce Spike Threshold Adaptive Learning in this study, which is designed to address the unique requirements of CLBP classification in the EmoPain dataset.

\section{Proposed Method} 
\label{sec:proposed_system}
This section details the proposed two-stage approach for classifying chronic lower back pain using biosignals. In the first stage, a novel learnable encoder -- Spike Threshold Adaptive Learning (STAL) -- converts biosignals (sEMG, Joint Angle, and Joint Energy) into spike trains. In the second stage, independent spiking recurrent neural networks are used within a multi-stream ensemble classification framework to analyse the encoded spike trains. This ensemble approach combines the predictions from each modality-specific SRNN using a Random Forest classifier to perform classification. Figure~\ref{fig:arch-overview} illustrates a schematic of the proposed architecture.

Let $\mathbf{X} \in \mathbb{R}^{T \times c}$ denote the input data, where $T$ is the number of time steps and $c$ is the number of channels. To maintain consistency in notation, we refer to the biosignal input dimensions as ``channels," irrespective of whether they represent physical sensor channels or derived features. The data is segmented into windows of size $\omega$ time steps, resulting in a collection of windowed data points denoted by $\mathbf{X}^{(\omega)} \in \mathbb{R}^{\omega \times c}$ for each window $\omega$.

\subsection{STAL: Spike Threshold Adaptive Learning} \label{sec:STAL}
\begin{figure}[!htbp]
    \centering
    \includegraphics[width=\linewidth]{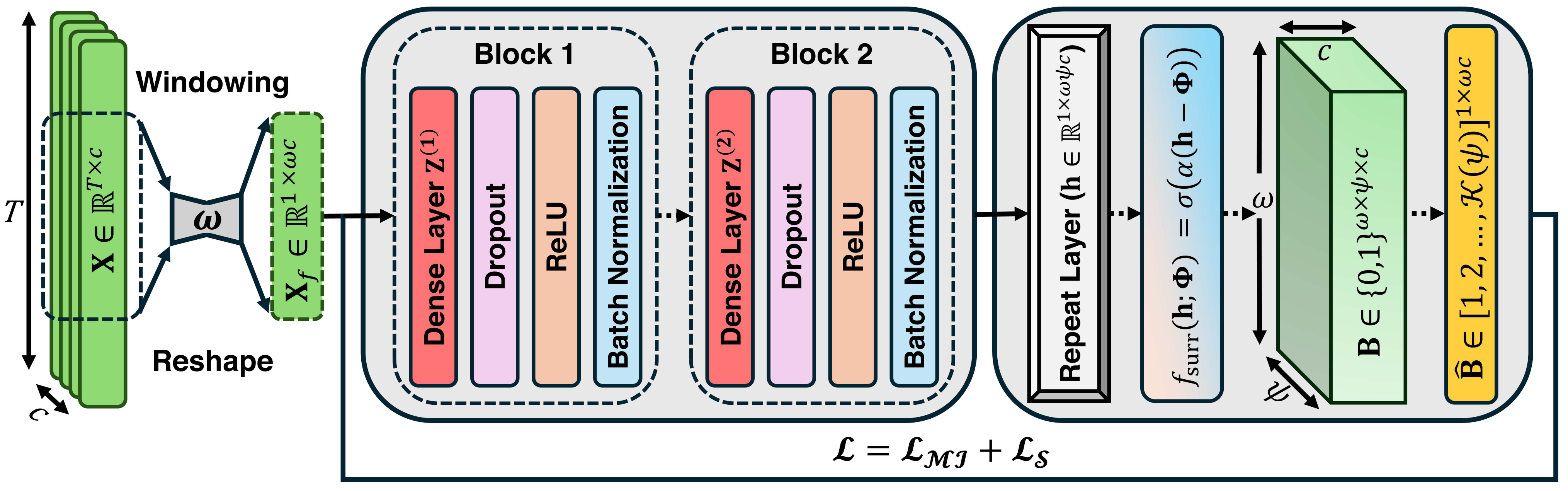}
    \caption{The STAL architecture consists of feature extraction and feature-to-spike conversion modules. Here, $\mathbf{\mathcal{L}}$ is the encoder loss function, formulated in Eq.~\eqref{eq:04}.}
    \label{fig:STAL}
\end{figure}

The STAL architecture, depicted in Fig.~\ref{fig:STAL}, consists of feature extraction and feature-to-spike conversion modules. The feature extraction module comprises two consecutive blocks, each containing a dense layer, dropout, ReLU activation, and batch normalisation. The ReLU activation function ($\mathrm{ReLU}(\cdot) = \max(0,\cdot)$) introduces non-linearity, allowing the network to learn complex patterns. Dropout helps prevent overfitting by randomly setting a fraction of input units to 0 during training. Batch normalisation improves the stability and performance of the neural network by normalising the inputs to each layer.

To prepare for feature extraction, we reshape $\mathbf{X}^{(\omega)}$ into a feature vector $\mathbf{X}_{f} \in \mathbb{R}^{1 \times \omega c}$. The first block processes $\mathbf{X}_{f}$ through a dense layer $\mathbf{Z}^{(1)}$, resulting in $\mathbf{Z}^{(1)} \in \mathbb{R}^{1 \times \ell}$, where $\ell$ denotes the number of neurons in the layer. The second block applies similar operations to the output of the first block, producing $\mathbf{Z}^{(2)} \in \mathbb{R}^{1 \times \ell}$.

The feature-to-spike conversion module of the STAL converts the extracted features into surrogate spiking activity. This process begins with an element-wise repetition operation on the output of the second block using Eq.~\eqref{eq:01}, expanding it to $\mathbf{h} \in \mathbb{R}^{1 \times \omega \psi c}$, where $\psi$ represents the number of spikes allowed at each time step.

\begin{equation} 
    \label{eq:01}
    \mathbf{h} = \mathrm{Repeat}(\mathbf{Z}^{(2)}, \psi)
\end{equation}

A learnable threshold vector, denoted as $\mathbf{\Phi} \triangleq \left [ \phi_{1}, \phi_{2}, \cdots, \phi_{\omega \psi c} \right ]^{\top}$, is incorporated into the encoding function $f: \mathbb{R} \rightarrow \{0,1\}$. This function is designed as a differentiable surrogate in Eq.~\eqref{eq:02} to generate binary spike trains $\mathbf{B} \in \{0,1\}^{\omega \times \psi \times c}$. Each threshold, denoted by $\phi_{i} \sim \mathcal{U}(a,b),~\forall i \in \{1,2,\cdots,\omega \psi c\}$, is initialised from a uniform distribution with hyperparameters $a, b \in \mathbb{R}^{+}$.

\begin{equation}   
    \label{eq:02}
    f_{\mathrm{surr}}(\mathbf{h}; \mathbf{\Phi}) = \sigma \left (\alpha \left (\mathbf{h} - \mathbf{\Phi} \right ) \right )
\end{equation}
where $\sigma (\cdot)$ is the sigmoid function and $\alpha$ controls the slope of the surrogate function. When $\alpha \rightarrow \infty$, the function approaches a true binary step function, mimicking the behaviour of real spiking neurons.

To align the dimensions of the spike train with the input for subsequent computations, we reshape $\mathbf{B}$ into $\mathbf{\hat{B}} \in \{1, 2, \cdots, \mathcal{K}(\psi)\}^{1 \times \omega c}$ using Eq.~\eqref{eq:03}. Here, $\mathcal{K}(\psi)$ denotes the number of possible combinations arising from summing the activations across $\psi$ spike positions at each time step and channel, capturing the temporal information about the order of spikes within $\mathbf{\hat{B}}$.

\begin{equation} 
    \label{eq:03}
    \mathbf{\hat{B}}_{i} = \sum_{j=1}^{\psi} p_{j} \mathbf{B}_{i,j,k}, \quad \forall i \in \{1,\cdots,\omega\}, k \in \{1,\cdots,c\}
\end{equation}
where $p_{j}$ represents the weight associated with the $j$-th spike position. This process incorporates the temporal information about the order of spikes into the final representation, as different spike orders will lead to different weighted sums and, consequently, different entries in $\mathbf{\hat{B}}$.

We introduce a customised loss function to update the parameters of the STAL encoder. This function consists of the Mutual Information term, inspired by~\cite{ferdaoussi2023information}, and the Sparsity term. The Mutual Information term ($\mathcal{L}_\mathcal{MI}$) measures the extent to which the encoder captures relevant information from the input data ($\mathbf{X}_{f}$) at multiple levels of representation, i.e., the encoded information in the hidden layers ($\mathbf{Z}^{(1)}$ and $\mathbf{Z}^{(2)}$) and encoded spike train ($\mathbf{\hat{B}}$). Meanwhile, the Sparsity term ($\mathcal{L}_{\mathcal{S}}$) serves as a regularisation term, encouraging the generation of sparse spike trains that efficiently represent the input data, promoting resource efficiency and robustness against overfitting. Formally, the loss function is formulated as in Eq.~\eqref{eq:04}.

\begin{align} 
    \label{eq:04}
    \mathcal{L}_\mathcal{MI} &= -\dfrac{1}{3}\left(\mathcal{I}\left[\mathbf{X}_{f}; \mathbf{Z}^{(1)}\right] + \mathcal{I}\left[\mathbf{X}_{f}; \mathbf{Z}^{(2)}\right] + \mathcal{I}\left[\mathbf{X}_{f}; \mathbf{\hat{B}} \right]\right), \nonumber \\
    \mathcal{L}_{\mathcal{S}} &= \lambda \cdot || \mathbf{X}_{f} - \mathbf{\hat{B}} ||_1, \nonumber \\
    \mathcal{L} &= \mathcal{L}_\mathcal{MI} + \mathcal{L}_\mathcal{S}
\end{align}
where $\mathcal{I}\left[\cdot;\cdot\right]$ computes the Mutual Information between tensors, $||\cdot||_1$ is the L1 norm, and $\lambda$ is the strength coefficient controlling the level of sparseness imposed on the internal representations.

To calculate Mutual Information, the encoded information in the hidden layers and spike train need to have the same dimensions as the input data. We reshape the hidden layer outputs when their dimensions differ. If $\ell < \omega c$, the features will be repeated $\left\lfloor \frac{\ell}{\omega c} \right\rfloor$ times. Conversely, if $\ell > \omega c$, average pooling will be used with a pool size and stride of $\left\lfloor \frac{\omega c}{\ell} \right\rfloor$.

The role of $\mathcal{L}_{\mathcal{S}}$ in regulating the sparsity of the spike train is essential. It uses L1 norm regularisation to calculate the difference between the input data ($\mathbf{X}_f$) and the surrogate spike train ($\mathbf{\hat{B}}$). This difference serves as an approximation of the number of spikes in the spike train. To limit the number of spikes to align approximately with the \emph{area} of the input signal, Eq.~\eqref{eq:05} defines the impact of the sparsity term on the training process. This control over sparsity is achieved by taking the derivative of the sparsity term with respect to $\mathbf{\hat{B}}$.

\begin{equation} 
    \label{eq:05}
    \dfrac{\partial \mathcal{L}_{\mathcal{S}}}{\partial \mathbf{\hat{B}}} = -\lambda \cdot \mathrm{sign}(\mathbf{X}_f - \mathbf{\hat{B}}).
\end{equation}
where $\mathrm{sign} (\cdot)$ is the sign function. This derivative encourages the output values to be close to 0 or 1, leading to a sparse output:

\begin{enumerate} 
    \item When $\mathbf{\hat{B}} > \mathbf{X}_f$, the derivative is negative, pushing output values towards 0.
    \item When $\mathbf{\hat{B}} < \mathbf{X}_f$, the derivative is positive, pushing output values towards 1.
    \item When $\mathbf{\hat{B}} = \mathbf{X}_f$, the derivative is 0, leaving output values unchanged.
\end{enumerate}

\subsection{Spiking Recurrent Neural Network} \label{sec:SRNN}
\begin{figure}[!htbp]
    \centering
    \includegraphics[width=0.9\linewidth]{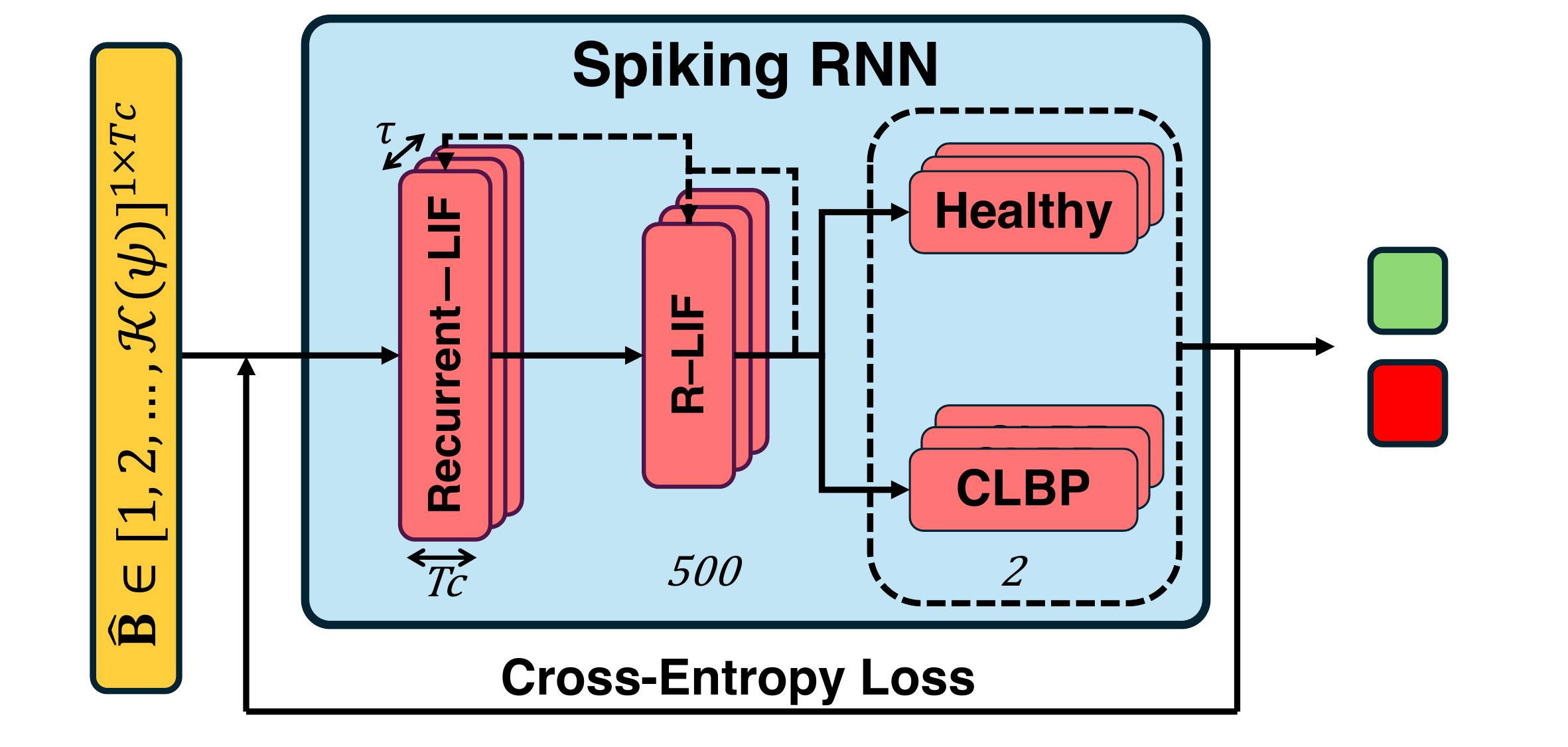}
    \caption{ The SRNN architecture. The spike trains $\mathbf{\hat{B}}$ of each data type (sEMG, Energy, Angle) are classified separately. $T$ represent the total time steps, $c$ are the channels, $\tau$ is the time dimension of the Recurrent Leaky Integrate-and-Fire (R-LIF) neurons.}
    \label{fig:SRNN}
\end{figure}

The proposed spiking recurrent neural network, depicted in Fig.~\ref{fig:SRNN}, leverages the temporal dynamics of spike trains for chronic lower back pain classification. The SRNN processes the output of the STAL encoder ($\mathbf{\hat{B}}$) through a network of Recurrent Leaky Integrate-and-Fire (R-LIF) neurons.

The network comprises $L$ layers of R-LIF neurons, where $L=2$ in our current implementation. The first layer (i.e., hidden layer) contains 500 neurons, and the second layer (i.e., readout layer) contains two neurons corresponding to the healthy and CLBP classes. All layers implement recurrent connections, allowing the network to capture complex temporal dependencies in the input spike trains.

Each R-LIF neuron is implemented using an exponentially decaying membrane potential, represented in discrete time steps. For each layer $l \in \{1, \cdots, L\}$, the membrane potential $\mathbf{U}^{l}[t]$ and output spikes $\mathbf{S}^{l}[t]$ at time step $t$ are computed by Eq.~\eqref{eq:06}.

\begin{align}
    \label{eq:06}
    \mathbf{U}^{l}[t+1] &= \beta\mathbf{U}^{l}[t] + \mathbf{I}^{l}[t+1] + \nonumber \\
    & \mathbf{V}^{l}(\mathbf{S}^{l}[t]) - \mathbf{R}^{l}[t] \mathbf{U}_{\mathrm{thr}}, \nonumber \\
    \mathbf{S}^{l}[t+1] &= \Theta(\mathbf{U}^{l}[t+1] - \mathbf{U}_{\mathrm{thr}}),
\end{align}
where $\beta = 0.99$ is the membrane potential decay rate chosen to enhance the LIF neurons' capacity to capture temporal dependencies in input spike trains by promoting longer memory. $\mathbf{I}^{l}[t]$ is the input current for layer $l$. The recurrent connections, $\mathbf{V}^l(\mathbf{S}^{l}[t])$, capture the impact of spikes from all interconnected neurons within layer $l$. The binary vector $\mathbf{S}^{l}[t]$ represents the output spikes of all neurons in layer $l$ at time $t$. We set the membrane threshold $\mathbf{U}_{\mathrm{thr}}$ to 1 in this study. The binary reset-by-subtraction mechanism $\mathbf{R}^{l}[t]$ is set to 1 when the threshold is reached and 0 otherwise, effectively subtracting $\mathbf{U}_{\mathrm{thr}}$ (which is 1 in this case) from the membrane potential. Lastly, $\Theta(\cdot)$ refers to the Heaviside step function.

All layers implement all-to-all connectivity. The input currents and recurrent connections for each layer are computed using Eq. \eqref{eq:07}.

\begin{align}
    \label{eq:07}
    \mathbf{I}^1[t] &= \mathbf{W}^{\{0,1\}}\mathbf{\hat{B}}[t], \nonumber \\
    \mathbf{I}^l[t] &= \mathbf{W}^{\{l-1,l\}}\mathbf{S}^{l-1}[t], \quad \text{for } l > 1, \nonumber \\
    \mathbf{V}^l(\mathbf{S}^l[t]) &= \mathbf{W}_{\mathrm{rec}}^l\mathbf{S}^{l}[t], \quad \text{for } l \in {1, \cdots, L},
\end{align}
where $\mathbf{W}^{\{0,1\}} \in \mathbb{R}^{n_1 \times c}$, $\mathbf{W}^{\{l-1,l\}} \in \mathbb{R}^{n_l \times n_{l-1}}$, and $\mathbf{W}_{\mathrm{rec}}^{l} \in \mathbb{R}^{n_l \times n_l}$ are learnable weight matrices. Here, $n_l$ represents the number of neurons in layer $l$, with $n_1 = 500$ and $n_2 = 2$ in our architecture, and $\mathbf{\hat{B}}[t]$ represents the input at time step $t$.

The SRNN processes the input over $\tau$ time steps, where $\tau=5$ is chosen to capture the temporal dynamics of the R-LIF neurons. The network's output is computed using Eq.~\eqref{eq:08}, which accumulates the spikes and membrane potentials of the output layer over all time steps.

\begin{align}
\label{eq:08}
    \mathbf{S}_{\mathrm{out}} &= \left[\mathbf{S}^{L}[1], \mathbf{S}^{L}[2], \cdots, \mathbf{S}^{L}[\tau]\right], \nonumber \\
    \mathbf{U}_{\mathrm{out}} &= \left[\mathbf{U}^{L}[1], \mathbf{U}^{L}[2], \cdots, \mathbf{U}^{L}[\tau]\right],
\end{align}
where the final classification is determined by the neuron with the highest cumulative activity across these $\tau$ time steps in the output layer.

To train the SRNN, we use the cross-entropy loss function, which is particularly suitable for our binary classification task between healthy and CLBP classes. The cross-entropy loss $\mathcal{L}$ is defined as in Eq.~\eqref{eq:09}.

\begin{equation}
    \label{eq:09}
    \mathcal{L}_\mathrm{SRNN} = -\sum_{i=1}^{C} y_i \log(\hat{y}_i)
\end{equation}
where $C$ is the number of classes (in our case, $C=2$), $y_i$ is the true label (0 or 1) for class $i$, and $\hat{y}_i$ is the predicted probability for class $i$. The predicted probabilities are obtained by applying a softmax function to the accumulated output $\mathbf{S}_{\mathrm{out}}$ and $\mathbf{U}_{\mathrm{out}}$. To handle the non-differentiable nature of spikes during backpropagation, we employ a surrogate gradient function. Specifically, we use the gradient of the shifted arc-tan function in Eq.~\eqref{eq:10}.

\begin{equation}
    \label{eq:10}
    \frac{\partial S}{\partial U} = \frac{1}{\pi} \frac{1}{\left( 1 + \left( \pi U \frac{\alpha}{2} \right)^2 \right)}.
\end{equation}

This surrogate gradient allows us to backpropagate the loss through the network and update the weights of the SRNN, enabling effective training despite the discrete nature of spike events.

\subsection{Multi-Stream Ensemble Classification} \label{sec:Ensemble}
We have implemented a multi-stream ensemble classification approach using separate STAL-SRNN pipelines to leverage the diverse information captured by various biosignal modalities. This tailored approach provides customised encoding and classification strategies for each type of data to compensate for the information loss during continuous-to-spike conversion and leverages cross-modal interactions. For each data modality $m \in \{\mathrm{sEMG}, \mathrm{Angle}, \mathrm{Energy}\}$, we train an independent STAL-SRNN pipeline formulated in Eq.~\eqref{eq:11}.

\begin{equation}
    \label{eq:11} 
    \mathbf{Y}_{m} = \mathcal{P}_{m}(\mathbf{\hat{B}}_{m}; \boldsymbol{\theta}_{m}), 
\end{equation}
where $\mathbf{\hat{B}}_{m}$ is the output of the STAL encoder for modality $m$, as defined in Eq.~\eqref{eq:03}. Here, $\mathcal{P}_m(\cdot)$ denotes the STAL-SRNN pipeline for that modality, $\boldsymbol{\theta}_{m}$ represents the learnable parameters of the pipeline, and $\mathbf{Y}_{m} \in \mathbb{R}^{2}$ is the output prediction (i.e., probabilities for Healthy and CLBP classes) for modality $m$. Each STAL-SRNN pipeline is trained independently using the cross-entropy loss function, as described in Section~\ref{sec:SRNN}. 

The outputs from these individual pipelines serve as features for the meta-classifier, which is implemented using a Random Forest classifier~\cite{randomforest}. This classifier is comprised of multiple decision trees, each trained on a bootstrap sample of the training data using the Gini impurity function to measure the quality of a split. The final prediction is determined by majority voting among the trees using Eq.~\eqref{eq:12}.

\begin{equation}
    \label{eq:12} 
    \hat{y} = \underset{C}{\mathrm{mode}} \left\{ \mathrm{RF}_{i} \left( \mathbf{Y}_{\mathrm{sEMG}}, \mathbf{Y}_{\mathrm{Angle}}, \mathbf{Y}_{\mathrm{Energy}} \right) \right\}_{i=1}^{N_t},
\end{equation}
where $\mathrm{RF}_{i}(\cdot)$ represents the prediction of the $i$-th decision tree in the Random Forest. The final prediction $\hat{y}$ is determined by majority voting among the $N_t$ trees.

\section{Experiments and Results} \label{sec:experiments_results}
\subsection{Dataset and Preparation}
The EmoPain dataset~\cite{aung2016EmoPain} is a multimodal collection of data from 46 participants, encompassing healthy individuals and those with CLBP. Participants performed various exercises (one-leg-stand, reach-forward, stand-to-sit, sit-to-stand, and bend-down) at normal and difficult levels, with transitions included to simulate real-life movements.

Data were collected using 18 wearable IMU sensors and four sEMG sensors. The IMU sensors recorded full-body 3D motion, from which 13 joint angle data (Joint Angle) and 13 angular energy data (Joint Energy) were derived. The sEMG sensors captured muscle activity data from the right and left upper and lower back muscle groups. The dataset includes self-reported pain intensity ratings (0-10) from CLBP individuals and labels for six pain-related behaviours provided by physiotherapists. We assigned positive labels to individuals reporting pain intensity greater than 5, resulting in 12 positive and 34 negative labels.

To prepare the data for analysis, we addressed the variation in recording lengths by trimming recordings that exceeded 18,000 frames by 6,000 frames from both ends. This method allowed us to balance standardisation with data preservation, reducing information loss and the impact of stress and fatigue~\cite{fernandez2023systematic}. We used normalisation techniques for all data modalities (sEMG, Angles, and Energies) to ensure consistent scaling and facilitate model training. Specifically, we normalised values to a 0-1 range by subtracting the minimum value and dividing by the maximum value within each sample. Additionally, we zero-padded the data to match the longest recording length (17,995 time steps). Lastly, we employed a sliding window approach with a 3,000 time-step window (50 seconds at 60 Hz), allowing the STAL to capture longer-term temporal dependencies and enhance robustness to noise.

\subsection{Implementation Details}
The experiments were conducted on a high-performance computing system equipped with an AMD EPYC 7302 16-core processor, 504GB of RAM, and an NVIDIA RTX A6000 GPU. The deep learning framework used was PyTorch version 2.3.1~\cite{torch}. The AdamW optimiser~\cite{adamw} was employed for both the STAL encoder and the SRNN classifier, with fixed learning rates of $5.0 \times 10^{-3}$ and $7.5 \times 10^{-4}$, respectively, due to its effectiveness in handling sparse gradients, a common characteristic in SNN training.

The number of neurons in the hidden layers was chosen based on the size of the sliding window. Therefore, both $\mathbf{Z^{(1)}}$ and $\mathbf{Z^{(2)}}$ were designed with 3000 neurons each to effectively process the temporal information within the data. The SRNN classifier's hidden layer was composed of 500 recurrent LIF neurons. This decision was deliberate to find a balance between computational efficiency, model performance, and generalisability. Additionally, to ensure that the model learns from all classes in the context of this imbalanced dataset with a small sample size and to mitigate any biasing effects, we applied minority over-sampling with weighted sample replacement during batch creation.

\begin{table*}[!htbp]
\centering
\caption{performance of four encoders on the EmoPain Database.}
\label{tab:results_emopain}
\begin{tabular}{llllllllll}
\hline
 &  &  &  &  &  & \multicolumn{4}{c}{Spike density $\downarrow$} \\ \cline{7-10} 
\multirow{-2}{*}{Encoder} & \multirow{-2}{*}{Classifier} & \multirow{-2}{*}{Accuracy $\uparrow$} & \multirow{-2}{*}{AUC $\uparrow$} & \multirow{-2}{*}{F1 $\uparrow$} & \multirow{-2}{*}{MCC $\uparrow$} & Ensemble & sEMG & Energy & Angle \\ \hline
\multicolumn{10}{l}{\cellcolor[HTML]{EFEFEF}Baseline methods} \\ \hline
Rate-coding~\cite{eshraghian2023snntorch} & SRNN & 73.91\% & 0.500 & 0.000 & 0.000 & 0.167 & 0.068 & 0.584 & 0.703 \\
Latency-coding~\cite{eshraghian2023snntorch} & SRNN & 71.74\% & 0.512 & 0.133 & 0.044 & 0.200 & 0.200 & \textbf{0.200} & \textbf{0.200} \\ \hline
\multicolumn{10}{l}{\cellcolor[HTML]{EFEFEF}Proposed method} \\ \hline
STAL--Stacked & SRNN & \textbf{80.43\%} & \textbf{0.679} & \textbf{0.526} & \textbf{0.437} & 0.549 & 0.360 & 0.703 & 0.793 \\
STAL--Vanilla & SRNN & 76.09\% & 0.596 & 0.353 & 0.270 & \textbf{0.021} & \textbf{0.007} & 0.439 & 0.528 \\ \hline
\end{tabular}
\end{table*}

\begin{figure}[!htbp]
    \centering
    \subfigure[]{\includegraphics[width=0.48\linewidth]{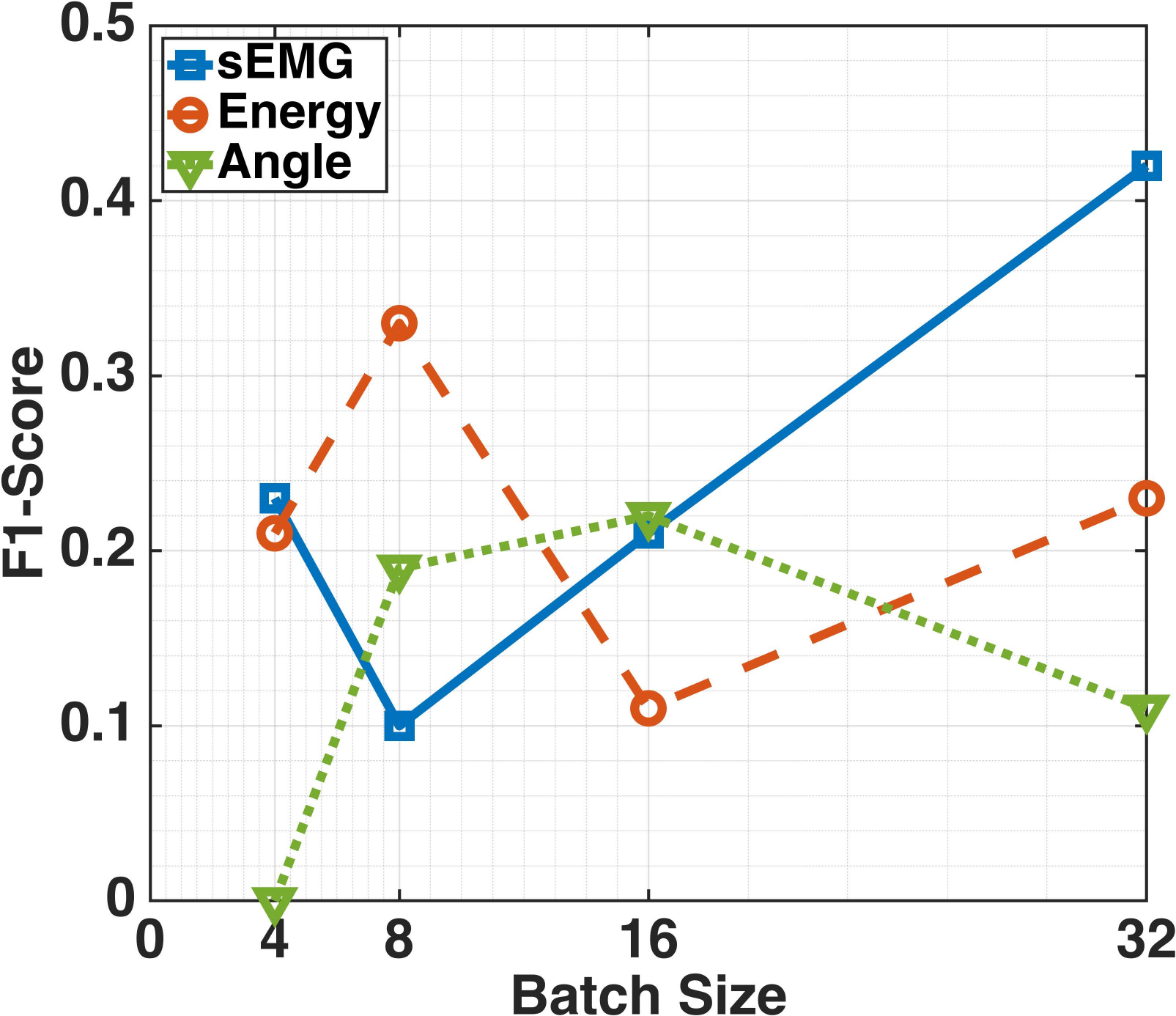}}
    \subfigure[]{\includegraphics[width=0.48\linewidth]{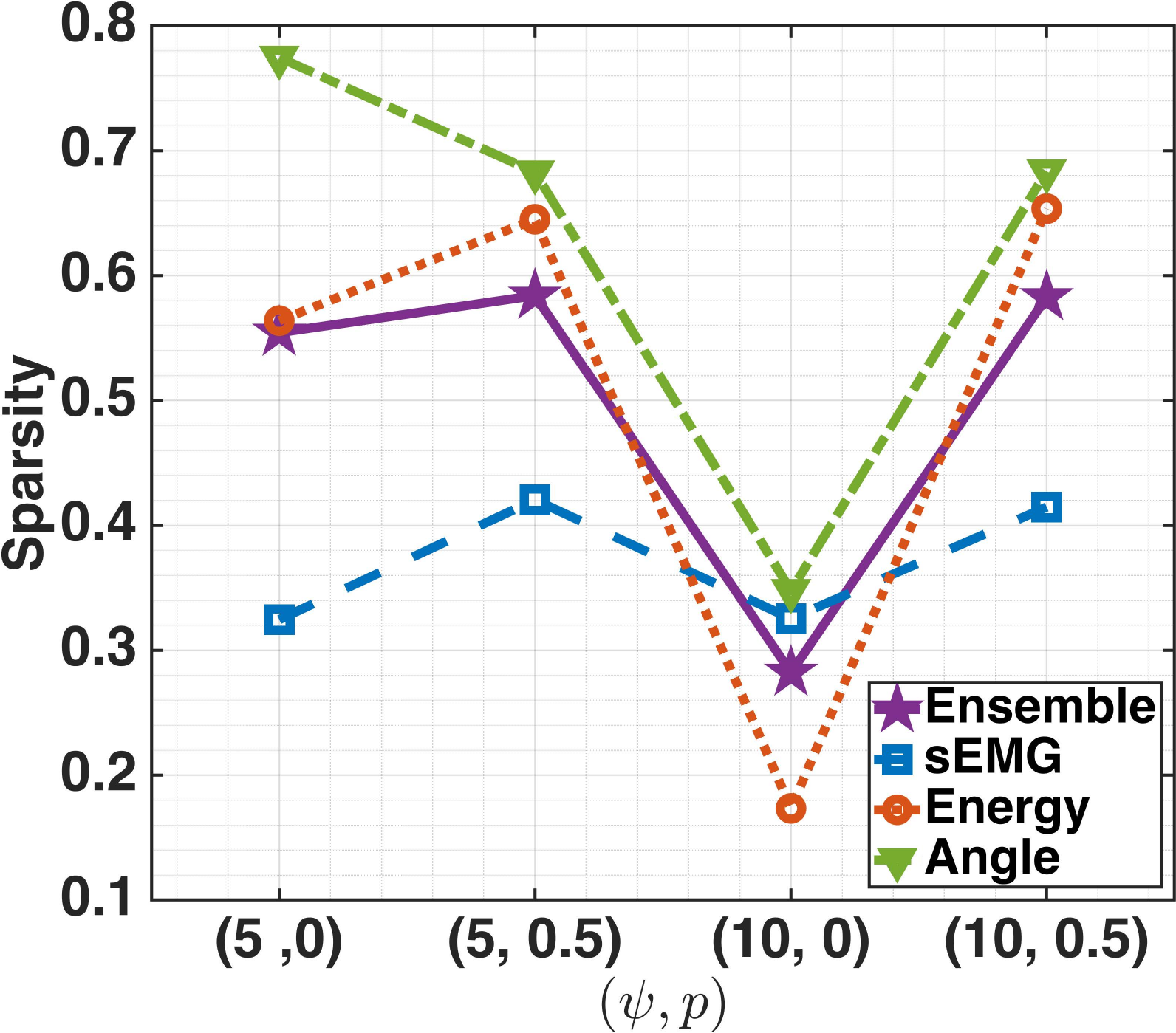}}
    \caption{(a) The influence of batch size on the F1-score of the proposed STAL-SRNN for each modality. (b) The Spike density for the proposed architecture as a function of $(\psi, p)$. While the difference in spike density between $(\psi, p) =(5,0.5)$ and $(\psi, p) =(10,0.5)$ is less than 1\% (0.549 vs 0.540), the former achieves better performance across other classification metrics.}
    \label{fig:parameter}
\end{figure}

An empirical study identified optimal hyperparameters for our STAL-SRNN architecture. The best configuration (see Fig.~\ref{fig:parameter}) included a batch size of 32 for sEMG, 8 for Energy, and 16 for Angle, along with a dropout rate of 0.5 and $\psi = 5$.

The training process consisted of 30 epochs for the STAL encoder and 25 epochs for the SRNN classifier, with early stopping implemented to prevent overfitting. This training regime was carefully chosen to balance model convergence and computational efficiency. All the hyperparameters and settings determined through this empirical search were subsequently used consistently across all experiments in this study. This approach ensures the reproducibility of our results and allows for a fair comparison across different experimental conditions\footnote{The implementation is accessible for reproducibility purposes on \url{https://github.com/freek1/emopain-stl}.}.

\subsection{Evaluation Metrics} \label{sec:eval}
We used both standard classification metrics and a sparsity measure tailored to SNNs to evaluate the effectiveness of the STAL-SRNN framework for CLBP classification on the EmoPain dataset. All metrics were assessed using leave-one-subject-out cross-validation to ensure a rigorous analysis, especially given the relatively small size of the EmoPain dataset.

The standard classification metrics employed in this study encompassed accuracy, Macro F1 score, Area Under the Curve (AUC), and Matthews Correlation Coefficient (MCC). These metrics were suggested in the EmoPain challenge~\cite{egede2020emopain} for comprehensive insights into the model's performance across various aspects.



In addition to these standard metrics, we introduced a spike density measure to assess the efficiency of our encoded spike trains. This measure provides insight into the sparsity of our neural representations across our multimodal approach. The overall spike density is computed as the harmonic mean of the individual densities for each data modality, as shown in Eq.~\eqref{eq:15}.

\begin{align}
    \label{eq:15}
    \mathrm{Spike Density} &= \frac{3}{\sum_{m} \frac{1}{D_{m}}}, \nonumber \\
    D_m &= \frac{\sum_{i=1}^{T}\sum_{j=1}^{\psi}\sum_{k=1}^{c_m} \mathbf{B}_{m,i,j,k}}{T \psi c_m},
\end{align}
where, $\mathbf{B}_m \in \{0,1\}^{T \times \psi \times c_m}$ represents the spike train for modality $m$, summed over $T$ time steps, $c_m$ channels, and $\psi$ spikes per time step. The denominator represents the maximum possible number of spikes for that modality.

\subsection{Experimental Results}
Our analysis focuses on three main aspects: (1) the performance of different spike train encoding strategies with the SRNN classifier, (2) an ablation study to understand the impact of architectural choices and hyperparameters, and (3) a comparison with state-of-the-art deep learning models for chronic pain assessment.

\subsubsection{Spike Train Encoders}
\begin{figure}[!htbp]
    \centering
    \includegraphics[width=\linewidth]{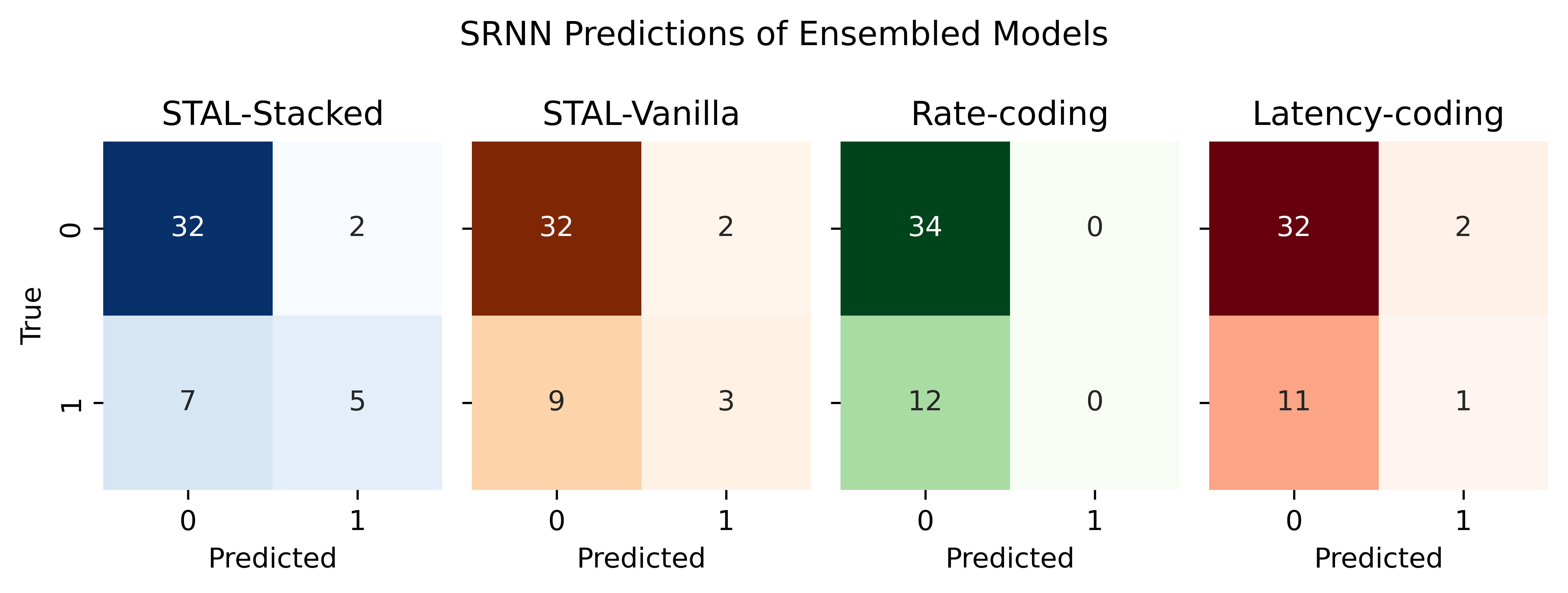}
    \caption{Confusion matrices of the encoders with the SRNN classifier.}
    \label{fig:cm-srnn}
\end{figure}

We conducted an assessment of four spike train encoders using the SRNN classifier with leave-one-subject-out cross-validation on the EmoPain dataset. The term STAL--Stacked refers to the proposed encoder outlined in Section~\ref{sec:STAL}. STAL--Vanilla represents a simplified version of STAL--Stacked, wherein the input is directly connected to the Repeat layer, excluding the two feature extraction blocks. Rate-coding and Latency-coding are considered baseline methods that exclusively encode information in the spatial and temporal domains, respectively. The experimental results are presented in Table~\ref{tab:results_emopain}, and Fig.~\ref{fig:cm-srnn} shows the confusion matrices for each encoder.

The results indicate that STAL--Stacked outperformed all other models, achieving the highest accuracy of 80.43\%, AUC of 0.679, F1 score of 0.526, and Matthews Correlation Coefficient (MCC) of 0.437. This suggests that STAL--Stacked offers the most robust and balanced performance among the encoders tested. STAL--Vanilla emerged as the second-best performer, achieving an accuracy of 76.09\%, AUC of 0.596, F1 score of 0.353, and MCC of 0.270. Notably, STAL--Vanilla achieved this with the least dense spike trains (spike density of 0.021), indicating its potential for efficient computation.

In comparison, rate-coding and latency-coding methods showed lower performance across all metrics. Rate-coding achieved 73.91\% accuracy but did not effectively differentiate between classes (F1 score and MCC of 0). Latency-coding performed slightly better in terms of F1 score and MCC but had the lowest overall accuracy at 71.74

\subsubsection{Ablation Study}
To assess the effectiveness of the STAL architecture and the impact of different design choices, we conducted an ablation study comparing STAL--Stacked and STAL--Vanilla and examining the effects of dropout probability.

The primary distinction between STAL--Stacked and STAL--Vanilla lies in the inclusion of feature extraction blocks. STAL--Stacked integrates two feature extractor blocks before the Repeat layer, while STAL--Vanilla directly connects the input to the Repeat layer. The superior performance of STAL--Stacked (80.43\% accuracy compared to 76.09\% for STAL--Vanilla) suggests that these additional layers contribute to a more effective feature representation. Additionally, STAL--Stacked employed a higher spike density (0.549) compared to STAL--Vanilla (0.021). This outcome indicates that the improved performance of STAL--Stacked is attributed to a more information-rich encoding, albeit at the cost of increased computational resources.

Both variants of the STAL method performed better than the baseline rate- and latency-based encoding approaches. This improvement can be attributed to STAL's unique ability to encode spike trains both spatially and temporally at the same time. Specifically, STAL encodes spatially by learning a distinct threshold value $\phi_{i} \in \mathbf{\Phi}$ for each feature $h_{i} \in \mathbf{h}$, as shown in Eq.~\eqref{eq:02}. At the same time, the loss function encourages temporal encoding of information by promoting spikes at the end of the spike train to encode a high signal, as demonstrated by the positional encoding in Eq.~\eqref{eq:03}. This dual encoding capability enables STAL to capture more complex patterns in the data, resulting in improved performance compared to the baseline methods.

In contrast, rate-coding encodes information in the spatial domain by proportionally correlating the probability sum of generated spikes for each input value while randomly positioning the spikes in the temporal domain. Similarly, latency-coding only encodes information temporally. This is evident from the uniform distribution in the spatial domain, where 1 spike is generated for each input, and the spike is strategically positioned in the temporal domain.
   
In our ablation study, we focused on examining the impact of dropout probability on the performance of our models. We found that in the STAL--Stacked architecture, a higher dropout probability $(p = 0.5)$ had a positive effect on performance. This can be attributed to the presence of fully connected layers in the feature extraction blocks, which facilitated information distribution during dropout. Conversely, in STAL--Vanilla, a lower dropout probability $(p = 0.0)$ was found to be optimal. This is because the absence of feature extraction layers in this architecture made it more susceptible to information loss during dropout.

\subsubsection{Comparison with Deep Learning Models}
To contextualise our results within the broader landscape of pain assessment methods, we compared our STAL-SRNN approach with state-of-the-art deep learning models on the EmoPain dataset (see Table~\ref{tab:sota}). This comparison serves multiple purposes:
\begin{enumerate}
    \item It establishes a performance baseline for spiking neural networks in chronic pain assessment, an area predominantly explored using traditional deep learning methods.
    \item It highlights the trade-offs between deep learning models' high performance and the potential efficiency advantages of spiking neural networks.
    \item It shows that despite the inherent differences in input processing (continuous vs. spike-based), our STAL-SRNN can achieve competitive results in certain aspects.
\end{enumerate}

\begin{table}[!htbp]
\centering
    \caption{Comparison of the proposed spiking architecture with state-of-the-art deep learning models on the EmoPain dataset.}
    \label{tab:sota}
    \begin{tabular}{llll}
    \hline
    Methods & AUC & F1 & MCC \\ \hline
    \multicolumn{4}{l}{\cellcolor[HTML]{EFEFEF}Deep Learning} \\ \hline
    MiMT~\cite{olugbade2021movement} & 0.648 & 0.722 & 0.327 \\
    LSTM+GCN~\cite{wang2021leveraging} & 0.690 & 0.731 & 0.365 \\
    GRU-RNN~\cite{dehshibi2023pain} & 0.855 & 0.765 & 0.411 \\
    L-SFAN~\cite{jorge2024LSFAN} & 0.849 & 0.772 & 0.510 \\ \hline
    \multicolumn{4}{l}{\cellcolor[HTML]{EFEFEF}Proposed Spiking Method} \\ \hline
    STAL-SRNN & 0.679 & 0.526 & 0.437 \\ \hline
    \end{tabular}
\end{table}

Our STAL-SRNN demonstrates competitive performance on the EmoPain dataset, achieving an AUC of 0.679 for CLBP classification. While this AUC is lower than the deep learning approaches (GRU-RNN: 0.855, L-SFAN: 0.849), it is important to highlight that our method outperforms the GRU-RNN model in MCC (0.437 vs 0.411). The competitive MCC suggests that STAL-SRNN has the potential for balanced classification in this domain.

\section{Conclusion and Future Work} \label{sec:conclusion}
In this study, we presented the first application of spiking neural networks for chronic lower back pain classification using the EmoPain dataset. Our work introduces two significant contributions to the field of neuromorphic computing and biosignal analysis. We introduced Spike Threshold Adaptive Learning (STAL), a novel, trainable encoder that effectively converts continuous biosignals into spike trains. Experimental results demonstrate that STAL could preserve crucial temporal dynamics and adapt to signal characteristics. We proposed an ensemble of Spiking Recurrent Neural Network (SRNN) classifiers designed to process surface Electromyography (sEMG) and Inertial Measurement Unit (IMU) data within a multi-stream architecture. This ensemble leverages a random forest classifier to efficiently harness the temporal processing capabilities inherent in SRNNs.

Our STAL-SRNN framework demonstrates outstanding performance, achieving an accuracy of 80.43\%, AUC of 67.90\%, F1 score of 52.60\%, and Matthews Correlation Coefficient (MCC) of 0.437. Notably, it outperforms traditional rate-based and latency-based encoding methods and surpasses the best-performing deep learning approach in terms of MCC, indicating better-balanced class prediction. This achievement was particularly significant given the challenges posed by the small sample size and class imbalance in the EmoPain dataset.

The success of our approach could represent a significant step forward in applying neuromorphic computing techniques to chronic pain management challenges. By leveraging the inherent efficiency of SNNs, our method bridged the gap between efficient spike-based computation and complex biosignal analysis. The STAL-SRNN provides new possibilities for personalised, continuous health monitoring and intervention in chronic pain management.

Future work could explore extending the framework to simultaneously classify pain levels and identify specific pain-related behaviours, providing more comprehensive insights for pain management. Furthermore, exploring the deployment of the STAL-SRNN framework on neuromorphic hardware could fully leverage its energy efficiency for long-term, wearable applications.

\bibliographystyle{IEEEtran}
\bibliography{reference}

\begin{thebibliography}{10}
\providecommand{\url}[1]{#1}
\csname url@samestyle\endcsname
\providecommand{\newblock}{\relax}
\providecommand{\bibinfo}[2]{#2}
\providecommand{\BIBentrySTDinterwordspacing}{\spaceskip=0pt\relax}
\providecommand{\BIBentryALTinterwordstretchfactor}{4}
\providecommand{\BIBentryALTinterwordspacing}{\spaceskip=\fontdimen2\font plus
\BIBentryALTinterwordstretchfactor\fontdimen3\font minus \fontdimen4\font\relax}
\providecommand{\BIBforeignlanguage}[2]{{%
\expandafter\ifx\csname l@#1\endcsname\relax
\typeout{** WARNING: IEEEtran.bst: No hyphenation pattern has been}%
\typeout{** loaded for the language `#1'. Using the pattern for}%
\typeout{** the default language instead.}%
\else
\language=\csname l@#1\endcsname
\fi
#2}}
\providecommand{\BIBdecl}{\relax}
\BIBdecl

\bibitem{supratak2016survey}
A.~Supratak, C.~Wu, H.~Dong, K.~Sun, and Y.~Guo, \emph{{Survey on Feature Extraction and Applications of Biosignals}}.\hskip 1em plus 0.5em minus 0.4em\relax {Springer International Publishing}, 2016, pp. 161--182.

\bibitem{li2024noninvasive}
N.~Li, R.~Zhou, B.~Krishna, A.~Pradhan, H.~Lee, J.~He, and N.~Jiang, ``{Non-invasive Techniques for Muscle Fatigue Monitoring: A Comprehensive Survey},'' \emph{{ACM Computing Survey}}, vol.~56, no.~9, 2024.

\bibitem{aung2016EmoPain}
M.~S.~H. Aung, S.~Kaltwang, B.~Romera-Paredes, B.~Martinez, A.~Singh, M.~Cella, M.~Valstar, H.~Meng, A.~Kemp, M.~Shafizadeh, A.~C. Elkins, N.~Kanakam, A.~De~Rothschild, N.~Tyler, P.~J. Watson, A.~C. D.~C. Williams, M.~Pantic, and N.~Bianchi-Berthouze, ``{The Automatic Detection of Chronic Pain-Related Expression: Requirements, Challenges and the Multimodal EmoPain Dataset},'' \emph{{IEEE Transactions on Affective Computing}}, vol.~7, no.~4, pp. 435--451, 2016.

\bibitem{yong2022prevalence}
R.~J. Yong, P.~M. Mullins, and N.~Bhattacharyya, ``Prevalence of chronic pain among adults in the united states,'' \emph{Pain}, vol. 163, no.~2, pp. e328--e332, 2022.

\bibitem{fernandez2023systematic}
R.~Fernandez~Rojas, N.~Brown, G.~Waddington, and R.~Goecke, ``A systematic review of neurophysiological sensing for the assessment of acute pain,'' \emph{{npj Digital Medicine}}, vol.~6, no.~1, p.~76, 2023.

\bibitem{makaram2021analysis}
N.~Makaram, P.~A. Karthick, and R.~Swaminathan, ``{Analysis of Dynamics of EMG Signal Variations in Fatiguing Contractions of Muscles Using Transition Network Approach},'' \emph{{IEEE Transactions on Instrumentation and Measurement}}, vol.~70, pp. 1--8, 2021.

\bibitem{dehshibi2023pain}
M.~M. Dehshibi, T.~Olugbade, F.~Diaz-de Maria, N.~Bianchi-Berthouze, and A.~Tajadura-Jiménez, ``{Pain Level and Pain-Related Behaviour Classification Using GRU-Based Sparsely-Connected RNNs},'' \emph{{IEEE Journal of Selected Topics in Signal Processing}}, vol.~17, no.~3, pp. 677--688, 2023.

\bibitem{farago2023review}
E.~Farago, D.~MacIsaac, M.~Suk, and A.~D.~C. Chan, ``{A Review of Techniques for Surface Electromyography Signal Quality Analysis},'' \emph{{IEEE Reviews in Biomedical Engineering}}, vol.~16, pp. 472--486, 2023.

\bibitem{jorge2024LSFAN}
J.~Ortigoso-Narro, F.~D. de~Maria, M.~M. Dehshibi, and A.~Tajadura-Jim\'{e}ne, ``{L-SFAN: Lightweight Spatially-focused Attention Network for Pain Behavior Detection},'' pp. 1--9, 2024.

\bibitem{wang2019acquisition}
F.~Wang, W.~M. Severa, and F.~Rothganger, ``{Acquisition and Representation of Spatio-Temporal Signals in Polychronizing Spiking Neural Networks},'' in \emph{{Proceedings of the 7th Annual Neuro-Inspired Computational Elements Workshop}}.\hskip 1em plus 0.5em minus 0.4em\relax Association for Computing Machinery, 2019.

\bibitem{sun2023feasibility}
A.~Sun, X.~Chen, M.~Xu, X.~Zhang, and X.~Chen, ``Feasibility study on the application of a spiking neural network in myoelectric control systems,'' \emph{Frontiers in Neuroscience}, vol.~17, p. 1174760, 2023.

\bibitem{auge2021reviewSNN}
D.~Auge, J.~Hille, E.~Mueller, and A.~Knoll, ``A survey of encoding techniques for signal processing in {Spiking Neural Networks},'' \emph{Neural Processing Letters}, vol.~53, no.~6, pp. 4693--4710, 2021.

\bibitem{perez2021sparse}
N.~Perez-Nieves and D.~F.~M. Goodman, ``{Sparse Spiking Gradient Descent},'' in \emph{{Advances in Neural Information Processing Systems}}, 2021, pp. 11\,795--11\,808.

\bibitem{SNN_POETS}
M.~Shahsavari, D.~Thomas, A.~Brown, and W.~Luk, ``{Neuromorphic Design Using Reward-based STDP Learning on Event-Based Reconfigurable Cluster Architecture},'' in \emph{{International Conference on Neuromorphic Systems}}.\hskip 1em plus 0.5em minus 0.4em\relax Association for Computing Machinery, 2021.

\bibitem{fabian2023sparse}
J.~M. Fabian, D.~C. O'Carrol, and S.~D. Wiederman, ``Sparse spike trains and the limitation of rate codes underlying rapid behaviours,'' \emph{{Biology Letters}}, vol.~19, no.~5, p. 20230099, 2023.

\bibitem{ARRAY2023}
M.~Shahsavari, D.~Thomas, M.~{van Gerven}, A.~Brown, and W.~Luk, ``{Advancements in spiking neural network communication and synchronization techniques for event-driven neuromorphic systems},'' \emph{{Array}}, vol.~20, p. 100323, 2023.

\bibitem{eshraghian2023snntorch}
J.~K. Eshraghian, M.~Ward, E.~O. Neftci, X.~Wang, G.~Lenz, G.~Dwivedi, M.~Bennamoun, D.~S. Jeong, and W.~D. Lu, ``{Training Spiking Neural Networks Using Lessons From Deep Learning},'' \emph{{Proceedings of the IEEE}}, vol. 111, no.~9, pp. 1016--1054, 2023.

\bibitem{donati2019discrimination}
E.~Donati, M.~Payvand, N.~Risi, R.~Krause, and G.~Indiveri, ``{Discrimination of EMG Signals Using a Neuromorphic Implementation of a Spiking Neural Network},'' \emph{{IEEE Transactions on Biomedical Circuits and Systems}}, vol.~13, no.~5, pp. 795--803, 2019.

\bibitem{gong2023spiking}
P.~Gong, P.~Wang, Y.~Zhou, and D.~Zhang, ``{A Spiking Neural Network With Adaptive Graph Convolution and LSTM for EEG-Based Brain-Computer Interfaces},'' \emph{{IEEE Transactions on Neural Systems and Rehabilitation Engineering}}, vol.~31, pp. 1440--1450, 2023.

\bibitem{gurchiek2021wearables}
R.~D. Gurchiek, N.~Donahue, N.~M. Fiorentino, and R.~S. McGinnis, ``{Wearables-Only Analysis of Muscle and Joint Mechanics: An EMG-Driven Approach},'' \emph{{IEEE Transactions on Biomedical Engineering}}, vol.~69, no.~2, pp. 580--589, 2021.

\bibitem{ferdaoussi2023information}
A.~El~Ferdaoussi, J.~Rouat, and E.~Plourde, ``Efficiency metrics for auditory neuromorphic spike encoding techniques using information theory,'' \emph{Neuromorphic Computing and Engineering}, vol.~3, no.~2, p. 024007, 2023.

\bibitem{randomforest}
T.~K. Ho, ``{Random decision forests},'' in \emph{{Proceedings of 3rd International Conference on Document Analysis and Recognition}}, vol.~1.\hskip 1em plus 0.5em minus 0.4em\relax IEEE, 1995, pp. 278--282.

\bibitem{torch}
A.~Paszke, S.~Gross, S.~Chintala, G.~Chanan, E.~Yang, Z.~DeVito, Z.~Lin, A.~Desmaison, L.~Antiga, and A.~Lerer, ``{Automatic differentiation in PyTorch},'' in \emph{NIPS Autodiff Workshop}, 2017.

\bibitem{adamw}
I.~Loshchilov and F.~Hutter, ``Decoupled weight decay regularization,'' in \emph{International Conference on Learning Representations}, 2019, pp. 1--18.

\bibitem{egede2020emopain}
J.~O. Egede, S.~Song, T.~A. Olugbade, C.~Wang, C.~D.~C. Amanda, H.~Meng, M.~Aung, N.~D. Lane, M.~Valstar, and N.~Bianchi-Berthouze, ``Emopain challenge 2020: Multimodal pain evaluation from facial and bodily expressions,'' in \emph{2020 15th IEEE International Conference on Automatic Face and Gesture Recognition (FG 2020)}.\hskip 1em plus 0.5em minus 0.4em\relax IEEE, 2020, pp. 849--856.

\bibitem{olugbade2021movement}
T.~Olugbade, N.~Gold, A.~C. d.~C. Williams, and N.~Bianchi-Berthouze, ``{A Movement in Multiple Time Neural Network for Automatic Detection of Pain Behaviour},'' in \emph{International Conference on Multimodal Interaction}.\hskip 1em plus 0.5em minus 0.4em\relax Association for Computing Machinery, 2021, pp. 442–--445.

\bibitem{wang2021leveraging}
C.~Wang, Y.~Gao, A.~Mathur, A.~C. De~C.~Williams, N.~D. Lane, and N.~Bianchi-Berthouze, ``{Leveraging Activity Recognition to Enable Protective Behavior Detection in Continuous Data},'' \emph{Proc. ACM Interact. Mob. Wearable Ubiquitous Technol.}, vol.~5, no.~2, 2021.

\end{thebibliography}

\end{document}